%% file: main.tex
\documentclass[lettersize,journal]{IEEEtran}
\pdfoutput=1
\usepackage{amsmath,amsfonts}
\usepackage{algorithmic}
\usepackage{algorithm}
\usepackage{array}
\usepackage[caption=false,font=normalsize,labelfont=sf,textfont=sf]{subfig}
\usepackage{textcomp}
\usepackage{stfloats}
\usepackage{url}
\usepackage{verbatim}
\usepackage{graphicx}
\usepackage{xcolor}
\usepackage{cite}
\usepackage{xcolor}
\usepackage{booktabs}
\usepackage{amssymb}
\usepackage{multirow}
\usepackage{pifont}
\usepackage{url}
\usepackage{hyperref}
\usepackage{tabularx}
\newcommand{\xmark}{\ding{55}}%
\def\BibTeX{{\rm B\kern-.05em{\sc i\kern-.025em b}\kern-.08em
    T\kern-.1667em\lower.7ex\hbox{E}\kern-.125emX}}

\begin{document}

\title{AICSD: Adaptive Inter-Class Similarity Distillation for Semantic Segmentation \\}

\author{Amir M. Mansourian, Rozhan Ahmadi, Shohreh Kasaei
        % <-this % stops a space
\thanks{The authors are with the Department of Computer Engineering, Sharif
University of Technology, Tehran 11155, Iran (e-mail: amir.mansurian@sharif.edu; roz.ahmadi@sharif.edu; kasaei@sharif.edu).}
}
%\markboth{MANUSCRIPT TO IEEE TRANSACTIONS ON INTELLIGENT TRANSPORTATION SYSTEMS}%
%{Shell \MakeLowercase{\textit{et al.}}: A Sample Article Using IEEEtran.cls for IEEE Journals}

\maketitle

\begin{abstract}
In recent years, deep neural networks have achieved remarkable accuracy in computer vision tasks. With inference time being a crucial factor, particularly in dense prediction tasks such as semantic segmentation, knowledge distillation has emerged as a successful technique for improving the accuracy of lightweight student networks. The existing methods often neglect the information in channels and among different classes. To overcome these limitations, this paper proposes a novel method called Inter-Class Similarity Distillation (ICSD) for the purpose of knowledge distillation. The proposed method transfers high-order relations from the teacher network to the student network by independently computing intra-class distributions for each class from network outputs. This is followed by calculating inter-class similarity matrices for distillation using KL divergence between distributions of each pair of classes. To further improve the effectiveness of the proposed method, an Adaptive Loss Weighting (ALW) training strategy is proposed. Unlike existing methods, the ALW strategy gradually reduces the influence of the teacher network towards the end of training process to account for errors in teacher's predictions. Extensive experiments conducted on two well-known datasets for semantic segmentation, Cityscapes and Pascal VOC 2012, validate the effectiveness of the proposed method in terms of mIoU and pixel accuracy. The proposed method outperforms most of existing knowledge distillation methods as demonstrated by both quantitative and qualitative evaluations. Code is available at: \textnormal{\href{https://github.com/AmirMansurian/AICSD}{https://github.com/AmirMansurian/AICSD}}
\end{abstract}

\begin{IEEEkeywords}
Deep Neural Networks, Semantic Segmentation, Knowledge Distillation, Inter-class Similarity, Intra-class Distribution, Adaptive Loss Weighting.
\end{IEEEkeywords}

\section{Introduction}
\input{docs/1_introduction}
%\input{temp/temp_introduction}

\section{RELATED WORK}
\input{docs/2_related_work}

\section{PROPOSED METHOD}
\input{docs/3_proposed_method}

% \input{tables/pascal-results}

% \input{tables/cityscapes-results}

\section{EXPERIMENTS}
\input{docs/4_experiments}

\section{CONCLUSION}
\input{docs/5_conclusion}

\section*{Acknowledgments}
The High Performance Center (HPC) of Sharif University of Technology is acknowledged by the authors for the provision of computational resources for this work.

\bibliographystyle{IEEEtran}
\bibliography{main}

\begin{IEEEbiography}[{\includegraphics[width=1in,height=1.25in,clip]{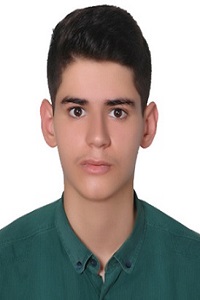}}]{Amir Mohammad Mansourian}
received the B.Sc. degree from the Department of
Electrical and Computer Engineering, Isfahan University of Technology, Isfahan, Iran, in
2021, and the M.Sc. degree from the Department of
Computer Engineering, Sharif University of Technology, Tehran, Iran, in 2023.\\
He is a member of Image Processing Laboratory (IPL) since 2021. His current research interests include Computer Vision, Image/Video Processing, and Deep Learning.
\end{IEEEbiography}

\begin{IEEEbiography}[{\includegraphics[width=1in,height=1.25in,clip]{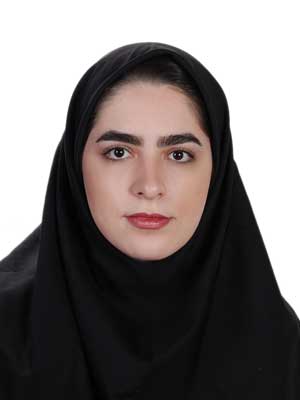}}]{Rozhan Ahmadi}
received the B.Sc. degree from the Department of 
Computer Engineering, Shahid Beheshti University, Tehran, Iran, in 2021, 
and the M.Sc. degree from the Department of
Computer Engineering, Sharif University of Technology, Tehran, Iran, in 2023. 
She is a member of Image Processing Laboratory (IPL) since 2021. Her current research interests include Computer Vision, Image/Video Processing, and Deep Learning.
\end{IEEEbiography}

\begin{IEEEbiography}[{\includegraphics[width=1in,height=1.25in,clip]{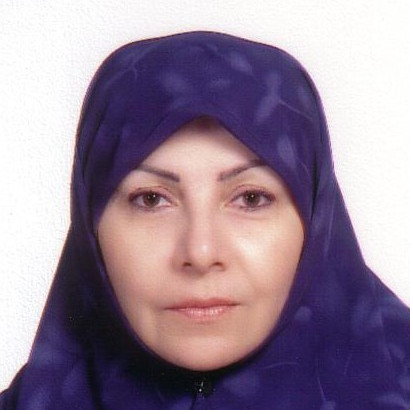}}]{Shohreh Kasaei}
(M’05–SM’07) received the B.Sc. degree from the Department of Electronics, the Department of Electrical and Computer Engineering, Isfahan University of Technology, Isfahan, Iran, in 1986, the M.Sc. degree from the Graduate School of Engineering, Department of Electrical and Electronic Engineering, University of the Ryukyus, Okinawa, Japan, in 1994, and the Ph.D. degree from Signal Processing Research Centre, School of Electrical and Electronic Systems Engineering, Queensland University of Technology, Queensland, Australia, in 1998.
\end{IEEEbiography}

\vspace{11pt}

\vfill

\end{document}

%% file: docs/1_introduction.tex
\IEEEPARstart{S}emantic Segmentation, as an essential element for understanding visual scenes, is a fundamental and challenging task in computer vision. It is a member in the group of dense prediction tasks with the objective of generating a labeling map, in which a particular class label is assigned to each pixel of the input image. Semantic segmentation has found numerous real-world applications in many fields; such as autonomous driving, video surveillance, scene and human-body parsing and many other areas.
\newline
Pioneered by the pivotal work of Fully Convolutional Network (FCN) \cite{long2015fully}, deep neural networks have significantly advanced the field of semantic segmentation. Since then, numerous FCN-based methods have been introduced to enhance segmentation performance by generating accurate segmentation maps. Those methods have improved segmentation accuracy through various approaches; such as employing stronger backbone networks \cite{ huang2017densely}, deeper network architectures with higher capacity compared to FCNs \cite{zhao2017pyramid}, incorporating multi-scale image contexts \cite{fu2019dual}, and refining segmentation details \cite{lin2016efficient}. These methods exhibit significant effectiveness in enhancing the performance and accuracy of semantic segmentation. However, the efficacy comes at the cost of efficiency; since the complexity in their model design requires substantial hardware resources, including significant computational power and memory requirements. As a result, their application in real-world is constrained, particularly on resource-limited devices such as mobile and other edge devices. In such real-world scenarios, the demand for lightweight and resource-efficient models becomes essential.
\input{figures/pull_figure} 
Given the mentioned concerns, lightweight neural networks with small model size and light computation cost have received significant attention. Quantization \cite{paszke2016enet, zhao2018icnet, mehta2018espnet}, pruning \cite{alvarez2016learning}, and decomposition \cite{he2016deep} of weights, present some of the approaches aiming to achieve lightweight networks. Those methods compress networks by using smaller precision for weights, removing redundant layers from the networks, and replacing large backbones with lighter versions, respectively. However, they have not fully closed the segmentation performance gap between compact networks and more complex ones.

Knowledge Distillation, first introduced by Bucila et al.~\cite{bucilua2006model} and popularized by~\cite{hinton2015distilling}, has served as a successful strategy for achieving a better trade-off between performance and efficiency of deep neural networks by using the knowledge of a more complex network (the teacher) to assist the training of a lighter network (the student). Methods based on knowledge distillation have greatly improved the accuracy of lightweight networks, performing tasks; such as image classification \cite{peng2019correlation, yue2020matching, chen2021distilling, ye2022generalized, chen2022knowledge}, object detection \cite{yang2022prediction, tang2022distilling, dai2021general}, and face recognition \cite{feng2020triplet, huang2022evaluation, li2023rethinking}. The knowledge distilled in the pioneering work of~\cite{ hinton2015distilling } for the task of image classification provided soft labels from a heavy teacher network with more beneficial information (e.g., intra-class similarity and inter-class difference), than the hard labels originally provided to the small network in the form of one-hot class label vectors. The fundamental idea of those methods is that soft labels offer the knowledge that hard ground truth labels are unable to convey. The student network is then supervised to mimic the teacher model using both hard and soft labels.

Knowledge Distillation has also been introduced to the field of semantic segmentation~\cite{ he2019knowledge, liu2019structured, shan2019distilling}. Liu et al.~\cite{ liu2019structured } viewed segmentation as the problem of classifying each pixel in an input image and, as a solution, applied knowledge distillation on pixel level (known as pixel-wise knowledge distillation). Semantic segmentation, being a dense prediction task, is used to predict dense structured outputs. While pixel-wise knowledge is effective for image classification, it is not enough to improve the performance of semantic segmentation. This is because aligning the pixel-level class distribution between teacher and student networks has the potential to disregard the contextual relationships among pixels and may even lead to performance degradation due to existing noise in activation maps. 

Considering these limitations, the adoption of complementary approaches such as pair-wise and holistic knowledge distillation has shown positive efficacy in enhancing the performance of semantic segmentation by capturing spatial structural context and complementary information beyond pixel-level knowledge. However, there are several constraints limiting the performance of these methods. To begin with, many prior works use the spatial relationships between feature maps or channels and the inter-class relations are ignored. Secondly, the negative effects of teacher on student’s training procedure are not thoroughly considered. Despite the progress made in defining different pair-wise distillation methods for transferring structured knowledge from teacher to student, existing methods do not consider inter-class similarities, which can be a valuable source of information for distillation. Moreover, as the teacher network itself has a lot of error in its predictions, it would not be reasonable to force the student to mimic the teacher's outputs throughout the training. Most existing works ignore this fact and set a constant scale for their distillation losses, and pixel-wise and pair-wise losses have a fixed impact on the training of the student network, which may lead to negative effects on the training process.

To address the aforementioned challenges, this paper proposes a pair-wise distillation method that leverages intra-class distributions and inter-class similarity matrices. The method defines class-specific intra-class distributions that capture the network's attention for each class and constructs inter-class similarity matrices that highlight similarities between these distributions. Figure \ref{fig:pull_figure} illustrates the intra-class distributions for a given image. Additionally, an adaptive loss weighting strategy is proposed that adjusts the scale of losses during student training process to emphasize the positive teacher knowledge and reduce the impact of negative information.

In summary, the main contributions of this work are as follows:

\begin{itemize}
\item Proposing a new pair-wise distillation method, called Inter-Class Similarity Distillation (ICSD), to transfer structured information from teacher to student.

\item Proposing a training strategy to control the impact of distillation in training phase of the student network by changing the scale of losses in an adaptive manner (ALW).

\item Validating the effectiveness of the proposed method with extensive experiments on the Cityscapes and PASCAL VOC 2012 datasets with state-of-the-art DeepLab V3+ segmentation network.  
\end{itemize}

%The proposed method can achieve high performance for semantic segmentation, establishing  \textbf{new state-of-the-art?} results on PASCAL VOC2012 (mIoU of \textbf{how much?}) and  Cityscapes (mIoU of \textbf{how much?}).

The remaining sections of this paper are structured as follows. Some related work relevant to the proposed method are reviewed in Section \ref{sec: re_work}. This is followed by a detailed explanation of the proposed method in Section \ref{sec: prop_method}. Lastly, extensive experiments and ablation studies are discussed in Section \ref{sec: experiments}. 
%Lastly, perspectives on future work are given in Section \ref{sec: conclusion}.

%% file: figures/pull_figure.tex
\begin{figure}[t]
\raggedleft
\centering
\includegraphics[scale=.4]{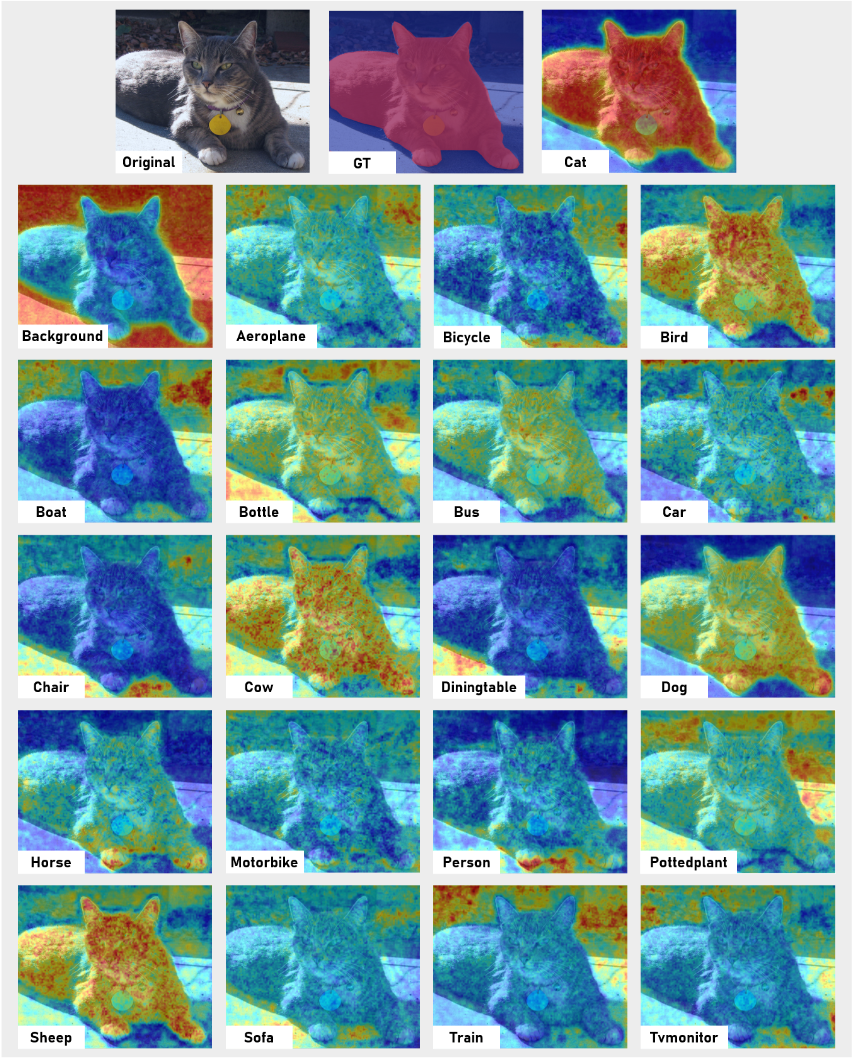}
\caption{Intra-class distributions for each class. Distributions are created by applying softmax to spatial dimension of output prediction of last layer. Similarities between each pair of intra-class distributions have good potential for distillation. Distributions are created from the PASCAL VOC 2012 dataset with 21 class categories.}
\label{fig:pull_figure}
\end{figure}

%% file: docs/2_related_work.tex
\label{sec: re_work}
In the following, the literature most relevant to this work is reviewed. This includes state-of-the-art research surrounding semantic segmentation and knowledge distillation.

\input{tables/related_work_comp}

\subsection{\textit{Semantic Segmentation}}
Semantic segmentation is a fundamental and challenging task in the field of computer vision. It allows for a deep, fine-grained understanding and analysis of the visual content. The problem is defined as assigning a particular class label to each pixel of an input image. The methods based on deep convolutional neural networks include pioneering work of FCN~\cite{long2015fully} followed by U-net~\cite{ronneberger2015u}, SegNet~\cite{badrinarayanan2017segnet}, and DeconvNet~\cite{noh2015learning}, in which managing to capture spatially structured context has been the key to their success.

Several approaches have been suggested in order to further improve the performance and accuracy. Incorporating boundary information~\cite{ding2019boundary} to better define object edges within the scene, extracting contextual information using  multi-context aggregation \cite{zhao2017pyramid} to capture spatial context in various scales from the image, enlarging the receptive field to gather more details regarding the long-rang relationships within pixels through the usage of multi-scale features and dilated convolutions  \cite{ chen2017deeplab, chen2018encoder, chen2017rethinking}, and attention-based methods \cite{fu2019dual, huang2019ccnet } to explore the connection between pixels and channels are some of the strategies incorporated to improve the semantic segmentation performance. This, however, comes with the cost of requiring high computational resources. The more complex the model design gets, larger and deeper networks are employed which demand considerable hardware and computational capabilities.

Various strategies have been proposed to make semantic segmentation models suitable for real-world applications, especially mobile applications. Some methods achieve higher efficiency by replacing heavy networks such as Deeplab-V3+\cite{chen2018encoder} and PSPNet\cite{zhao2017pyramid} with lighter models where the encoder module is less complex; e.g., Mobilenet-V2 \cite{sandler2018inverted} and Resnet18 \cite{he2016deep}. Other approaches have proposed lowering the cost of convolutional operations to increase efficiency.  Enet \cite{paszke2016enet} uses filter factorization,  lighter encoder/decorder, and early downsampling to create an efficient network. ESPNet \cite{mehta2018espnet} replaces standard convolutions with a module that is a combination of efficient spatial pyramid of convolutions and point-wise convolutions. ICNet \cite{zhao2018icnet} is an image cascade network where features from low and high resolution images are fused to maintain a balance between efficiency and accuracy. BiSeNet \cite{yu2018bisenet} achieves this goal by using the combination of a spatial and contextual path to increase feature processing efficiency.

\subsection{\textit{Knowledge Distillation}}
Knowledge Distillation is another common approach used in semantic segmentation to balance the trade-off between accuracy and efficiency. Introduced by Bucila et al. \cite{bucilua2006model} and popularized by \cite{ hinton2015distilling }, knowledge distillation trains a smaller and compact network called the student, using the knowledge distilled from a heavier cumbersome network called the teacher. The fundamental approach is to minimize the KL-divergence between
the logits outputed by the teacher and student by getting the student to imitate the actions of the teacher. These logits, considered as soft labels, provide the student network with substantial information from the cumbersome teacher network that the original hard labels (i.e., one-hot class label vectors) are unable to collect. 

Other methods have explored the distilling knowledge based on corresponding features across teacher and student networks. FitNet \cite{ romero2014fitnets } aligns feature maps extracted from intermediate hidden layers. RKD \cite{ park2019relational } distills distance and angle wise correlations between features. In AT \cite{ zagoruyko2016paying }, the student is trained to imitate the corresponding intermediate attention map from teacher. \cite{zhou2020channel} calculates weights for each channel of a feature map, which is the importance of each channel, and then distills these weights. Also, decreases the impact of distillation loss with the increase of epochs. \cite{ huang2017like} employs the Gram Matrix \cite{ gatys2016image} to distill the correlation between feature maps activated and selected considering the distributions of neuron selectivity patterns. These strategies have managed to improve the performance of lightweight networks without causing any increase in their inference load, in many cases even making these student networks considerably faster. 

\subsection{\textit{Knowledge Distillation for Semantic Segmentation}}
Knowledge distillation has been employed in creating fast and compact semantic segmentation networks \cite{heo2019comprehensive, wang2020intra, shu2021channel, an2022efficient, yang2022cross}. SKDS \cite{ liu2019structured } applied knowledge distillation in pixel-level. Its strategy, however successful in other tasks, is limited in improving the performance of semantic segmentation. The focus on individual pixel-wise relations in these methods, fail to capture the structural context needed in dense prediction. Taking these restrictions into account, SKDS  has also demonstrated that incorporating complementary strategies (i.e., pair-wise and holistic knowledge distillation alongside pixel-wise knowledge distillation) can be effective in improving the performance of semantic segmentation by taking cross-pixel relationship dependencies into consideration. Holistic knowledge distillation aligns high-order relations between the segmentation maps distilled from the teacher into the student network.

\input{figures/method_diagram}
This work is mostly focused on pair-wise knowledge distillation. Based on pair-wise Markov random field framework \cite{ li2009markov }, SKDS proposes to incorporate pair-wise similarity relations among pixels to improve spatial labeling contiguity. Xie et al. \cite{ xie2018improving} have also employed a pixel-level feature map-based pair-wise distillation method by capturing local pixel probability differences among each pixel and its eight neighbours. Motivated by that, several subsequent proposed studies have achieved notable improvements. Feng et al. \cite{ feng2021double } distill class-level pair-wise relationships through similarity within category dimensions. Liu et al. \cite{ liu2021exploring } capture the similarities and variations across the channels of a feature map. Yim et al. \cite{ yim2017gift} defined the flow between network layers by calculating the inner product between features maps from consecutive pair of levels. Tung et al. \cite{ tung2019similarity} consider instance-level pair-wise distillation. This is done by computing the (dis)similarities between samples in a batch and train the student network to preserve the relations in its own representation space. Channel-level pair-wise distillation is explored in \cite{ park2020knowledge, liu2021exploring}. Park et al. \cite{ park2020knowledge } propose a channel and spatial correlation (CSC) loss to extract and transfer the full long-range relationship in the feature map. Table \ref{tab: related_work_comp} shows a summary of pair-wise losses for knowledge distillation implemented in the mentioned methods.

In this study, a pair-wise distillation method is proposed, similar to \cite{liu2021exploring}. The method involves computing intra-class distributions for each class, followed by the calculation of an inter-class similarity matrix for distillation purposes. Additionally, an adaptive training strategy for the student network using a distillation loss is investigated, drawing inspiration from \cite{zhou2020channel}.

%% file: tables/related_work_comp.tex
\begin{table*}[ht]
\caption{Loss Implementation in Existing Pair-wise Knowledge Distillation Methods.}
\centering
\begin{tabular}{c c c c c}
\toprule
\textbf{Method} & \textbf{Alignment Order} & \textbf{Distance} & \textbf{Formulation} & \textbf{Information} \\ \midrule
MD \cite{xie2018improving}   & Pixel Level      & L2 & $L_{2}(y,x)$               & 8-Neighbourhood Consistence Map           \\ \midrule
SKD \cite{liu2019structured} & Pixel Level      & L2 & $\frac{f_{i}^{T}f_{j}}{(\left \| f_{i} \right \|_{2}\left \| f_{j} \right \|_{2})}$ & Pixel Similarity                         \\ \midrule
CSC \cite{park2020knowledge} & Channel Level    & L2 & $\bigoplus_{c=1}^{C} f_c(i,j) \otimes f_{s_{_{c}}}(i,j)$ & Set of Channel Correlations              \\ \midrule
ICKD \cite{liu2021exploring} & Channel Level    & L2 & $G^{F^{^{T}}} = f(F^{T})\cdot f(F^{T})^{T}$ & Inter-Channel Correlation Matrix         \\ \midrule
CSD \cite{feng2021double}    & Class Level      & L2 & $CM(q)_{ij} = N(q^{i.})\cdot N(q^{j.})$ & Cross-Category Correlation Matrix        \\ \midrule
FSP \cite{yim2017gift}       & Feature map Level & L2 & $\sum_{s=1}^{h}\sum_{t=1}^{w}\frac{F_{s,t,i}^{1}(x;W)\times F_{s,t,j}^{2}(x;W)}{h\times w}$ & Cross-Layer FSP Matrix                   \\ \midrule
SP \cite{tung2019similarity} & Instance Level   & L2 & $\frac{\tilde{G}_{T}^{(l)}}{\left \| \tilde{G}_{T}^{(l)} \right \|_{2}}$ & Cross-Instance   \\ \bottomrule                      
\end{tabular}
\label{tab: related_work_comp}
\end{table*}

%% file: figures/method_diagram.tex
\begin{figure*}[!ht]
\centerline{\includegraphics[scale=.21]{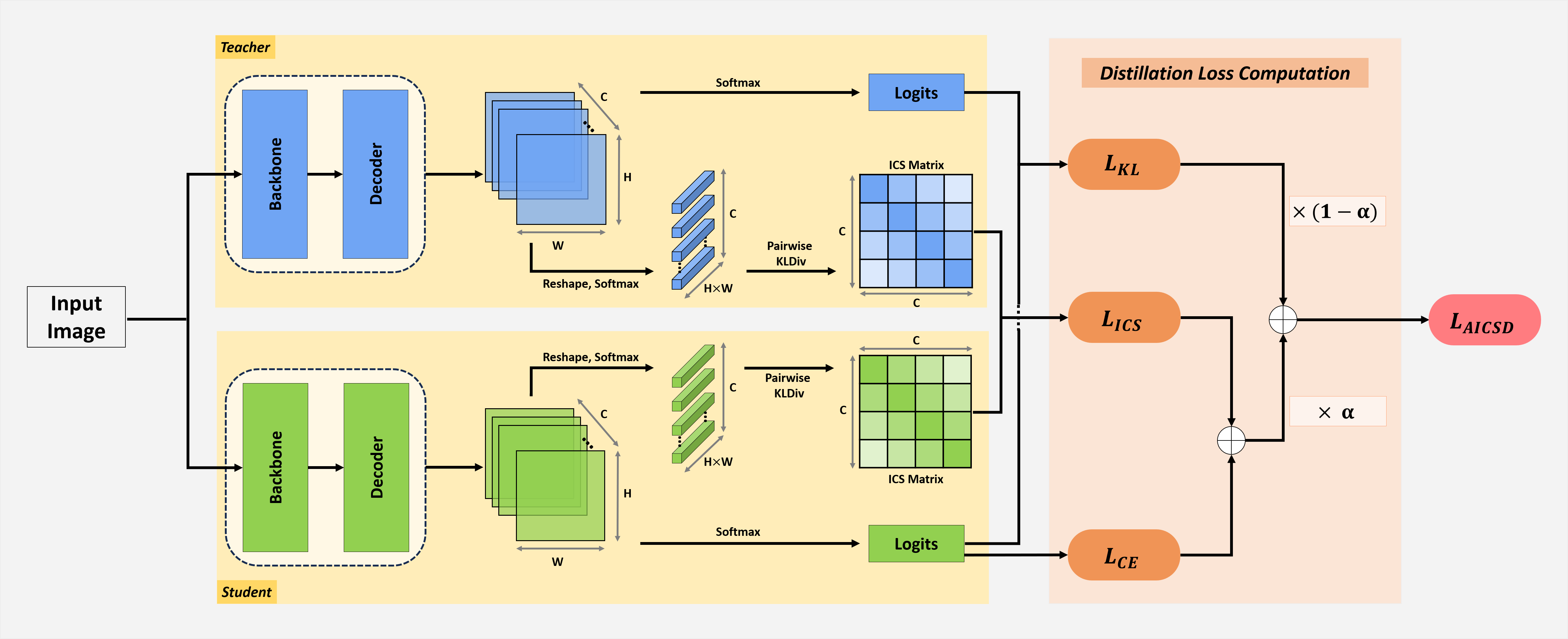}}
\caption{\textbf{Overall diagram of the proposed AICSD}. Network outputs are flattened into 1D vectors, followed by application of a softmax function to create intra-class distributions. KL divergence is then calculated between each distribution to create inter-class similarity matrices. An MSE loss function is then defined between the ICS matrices of the teacher and student. Also, KL divergence is calculated between the logits of the teacher and student for pixel-wise distillation. To mitigate the negative effects of teacher network, an adaptive weighting loss strategy is used to scale two distillation losses and cross-entropy loss of semantic segmentation. During training, hyperparameter $\alpha$ undergoes adaptive changes and progressively increases with epoch number.}
\label{fig:method_diagram}
\end{figure*}

%% file: docs/3_proposed_method.tex
\label{sec: prop_method}
In this section, first, the basic knowledge distillation technique\cite{hinton2015distilling} is reviewed. Subsequently, the proposed inter-class similarity distillation method is explained. Finally, the training strategy for adding distillation losses to the Cross Entropy (CE) loss of semantic segmentation is examined. The method diagram of the proposed approach is shown in Figure \ref{fig:method_diagram}.

\subsection{Preliminary}
Let $Z^T, Z^S \in \mathbb{R} ^{C\times H\times W}$ denote the outputs of the last layer of the teacher (T) and student (S) networks, respectively. Notation $Z_k(i,j)$ is used to show the element at depth $k$, and spatial dimensions are indexed by $i$ and $j$. The first knowledge distillation method, originally proposed by Hinton et al.\cite{hinton2015distilling}, aimed to use the class probabilities generated by the teacher network as soft labels for training the student network. This was done by minimizing the Kullback-Leibler (KL) divergence between the class probabilities of the teacher and student networks. In the context of semantic segmentation, as a collection of separate classification tasks, the distillation loss can be written as

\input{equations/vanilla_kd}
where $\sigma(.)$ and $\tau$ denote the softmax function and temperature factor, respectively, and $KL(.||.)$ represents the KL divergence between the logits of the T and S networks. In this work,  this distillation method is implemented in our own method, called KD method, and it's loss is written as $\ell_{KD}$.

\subsection{Inter-Class Similarity Distillation}
In this section, the formulations for creating an inter-class similarity matrix from intra-class distributions are discussed in detail. Given $Z^T$ and $Z^S$ as the outputs of the teacher and student networks for a specific image, intra-class distributions are created by applying softmax at the spatial dimension of $Z^t$ and $Z^s$. This leads to intra-class information for each class, regardless of other classes, and a probability distribution is provided for a specific class. Considering $G \in \mathbb{R} ^{C\times HW}$ as intra-class distributions, the formulations can be written as 

\input{equations/intra-class_distributions}

where Flatten(.) function vectorizes a 2D output map ($H \times W$) into a 1D vector with length $HW$. Then, the Inter-Class Similarity (ICS) matrix can be created from the intra-class distributions by calculating the KL divergence between each pair of intra-class distributions as

\input{equations/ICS_matrix}
Finally, the inter-class similarity loss between teacher and student is defined by

\input{equations/ICS_loss}
where $ICS^T$ and $ICS^S$ denote ICS matrices of teacher and student and $C$ is the number of classes. As last layer outputs of the both teacher and student networks have the same number of channels and spatial dimension, there is no need to change the spatial size or number of channels of student to match the teacher network and the proposed pair-wise distillation method can be applied on any segmentation network. 

%As our method is applied to the last layer outputs of the network, there is no need to change the spatial size or number of channels of student to match the teacher network.

This distillation method is called Inter-Class Similarity Distillation (ICSD), and its final objective is given by
\input{equations/ICS_total_loss}
where $\ell_{CE}$ denotes the conventional Cross Entropy (CE) loss of semantic segmentation task and $\lambda$ is a hyperparameter that controls the scale of proposed distillation loss.

\subsection{Adaptive Loss Weighting}
Although distillation improves the accuracy of the student, the teacher network itself can still have error in its predictions. \cite{cho2019efficacy} investigated the negative impacts of distillation and \cite{zhou2020channel} proposed an early decay strategy to decrease the effect of their channel-wise distillation loss in the last epochs of training phase. Inspired by those methods, an Adaptive Loss Weighting (ALW) process is utilized to mitigate the negative impacts of distillation. In this strategy, the standard distillation method is used to train the student network in the first epochs, as the soft logits provided by the teacher are easier to learn from than zero-one labels. However, towards the end of training phase, the focus shifts to the CE loss and ICSD loss, which transfer the structure of the teacher to the student, allowing the student to take control of the training process. The final loss of the proposed method with the ALW strategy is defined as
\input{equations/final_loss}
where the hyperparameter $\alpha$ controls the weighting of losses according to the ALW strategy and can be adjusted either linearly
\input{equations/ALW_lin}
where $e$ is the epoch number and $N_e$ is the number of all epochs, or adjusted exponentially 
\input{equations/ALW_expo}
where $\beta$ is a hyperparameter.

In fact, this process is, in general, analogous to the human learning system in which a child, in early stages, cannot learn on its own without the guidance of a teacher. As the student network becomes more trained, it can gradually learn on its own, by using labels, with only a structural guidance (pair-wise loss) from the teacher network. This training strategy can be used in any other distillation methods with pixel-wise or pair-wise distillation losses.

%where $e$ denotes epoch number and $N_e$ is the number of all epochs in training process.

%% file: equations/vanilla_kd.tex
\begin{equation} \label{eq: vanilla_kd} 
    \centering
\ell_{kd} = \frac{1}{H \times W} \sum_{i=1}^{H} \sum_{j=1}^{W} KL(\sigma(\frac{z^T(i,j)}{\tau})||(\sigma(\frac{z^S(i,j)}{\tau})),
\end{equation}

%% file: equations/intra-class_distributions.tex
\begin{equation} \label{eq: intra-class_distributions} 
    \centering
G_i = \sigma(Flatten(Z_i)),
\end{equation}

%% file: equations/ICS_matrix.tex
\begin{equation} \label{eq: ICS_matrix} 
    \centering
ICS(i, j) = KL(G_i || G_j).
\end{equation}

%% file: equations/ICS_loss.tex
\begin{equation} \label{eq: ICS_loss} 
    \centering
\ell_{ICS} = \frac{1}{C^2} \parallel ICS^T - ICS^S \parallel_2^2,
\end{equation}

%% file: equations/ICS_total_loss.tex
\begin{equation} \label{eq: ICS_total_loss} 
    \centering
\ell_{total} = \ell_{CE} + \lambda \ell_{ICSD}
\end{equation}

%% file: equations/final_loss.tex
\begin{equation} \label{eq: final_loss} 
    \centering
\ell_{AICSD} = \alpha (\ell_{CE} + \ell_{ICSD}) + (1-\alpha) \ell_{kd},
\end{equation}

%% file: equations/ALW_lin.tex
\begin{equation} \label{eq: Linear-ALW} 
    \centering
 \alpha = \frac{e-1}{N_e},
\end{equation}

%% file: equations/ALW_expo.tex
\begin{equation} \label{eq: exp-ALW} 
    \centering
 \alpha = \beta ^ {e-1},
\end{equation}

%% file: docs/4_experiments.tex
\label{sec: experiments}
This section begins by introducing the datasets, evaluation metrics, and implementation details. Next, the results of the proposed method are reported and compared with those of some existing distillation methods. Finally, ablation studies are discussed to more validate the proposed method.

\subsection{Datasets}

\subsubsection{\textit{Pascal VOC 2012}}
The Pascal VOC dataset is a widely used computer vision dataset for object recognition and segmentation, containing 1,464 labeled images for training, 1,449 for validation, and 1,456 for testing, with 21 foreground object categories including background class. In this work, an augmented version of the dataset is used that includes extra annotations, as provided by \cite{chen2018encoder}.

\subsubsection{\textit{Cityscapes}}
The Cityscapes dataset was created for the purpose of urban scene understanding and includes 30 object classes, although only 19 of these classes are used for evaluation purposes. The dataset consists of 5,000 high-quality images that have been finely annotated at the pixel level, as well as an additional 20,000 images that have been coarsely annotated. The finely annotated images are divided into three sets: 2,975 for training, 500 for validation, and 1,525 for testing. In this work, only the subset of 5,000 finely annotated images is used.

\input{figures/miou_classwise}
\subsection{Evaluation Metrics}
The segmentation performance is evaluated using two metrics: mean Intersection-over-Union (mIoU) and pixel accuracy. The IoU metric measures the ratio of intersection to union between the predicted results and ground truth for each object category, and the mIoU is the average of the IoUs across all categories. Additionally, the pixel accuracy metric measures the ratio of correctly predicted pixels to all the pixels. To represent the model size, the number of network parameters is reported.
\input{tables/pascal-results}
\input{tables/cityscapes-results}

\subsection{Implementation Details}
\subsubsection{\textit{Network Architectures}}
For a fair evaluation, the experiments are conducted using the same teacher and student network as in \cite{liu2021exploring}. Specifically, the teacher network used in all of the experiments is Deeplab V3+ with ResNet101 as the backbone. For the student network, Deeplab V3+ segmentation with different backbones including ResNet18 and MobileNetv2 is used. 

\subsubsection{\textit{Training Details}}
The student networks are trained using a similar configuration, which includes a batch size of 6 and total number of 120 epochs for pascal dataset and batch size of 4 and total number of 50 epochs for the Cityscapes dataset. The stochastic gradient descent (SGD) optimizer with an initial learning rate of 0.007 (pascal), and 0.01 (cityscapes) is used and it is reduced according to the cosine annealing scheduler. Prior to training, each image is preprocessed using random scaling to 0.5 to 2 times of their original size, horizontal random flipping, and a random crop of $513\times513$ pixels for pascal dataset, and $512\times1024$ for the Cityscapes dataset. The teacher and student networks use pre-trained weights from the ImageNet dataset for their backbones, while their segmentation parts are initialized randomly. The hyperparameters defined in the equations \ref{eq: ICS_total_loss} and \ref{eq: exp-ALW} were fine-tuned by testing different values and selecting the optimal ones. After experimentation, the values that produced the best results were $\lambda$=9500 and $\beta$=0.985. For inference, the performance is evaluated on a single scale and original inputs and results are average of three runs. The implementation is done using the PyTorch framework. All networks are trained on a single NVIDIA GeForce RTX 3090 GPU.

\subsection{Experimental Results}
To evaluate the performance of the proposed pair-wise method, extensive experiments were conducted and compared to several existing distillation methods including: KD \cite{hinton2015distilling}, Attention Transfer (AT) \cite{zagoruyko2016paying}, Similarity preserving (SP) \cite{tung2019similarity} and Inter-channel Correlation Knowledge Distillation (ICKD) \cite{liu2021exploring}. %which are listed below:

\begin{comment}
\begin{itemize}
\item Known knowledge Distillation (KD)\cite{hinton2015distilling}: This method uses the teacher's outputs as soft labels to transfer knowledge to the student network by minimizing the Kullback-Leibler divergence between the logits of the teacher and student networks.\\

\item Attention Distillation (AD)\cite{zagoruyko2016paying}: A pixelwise loss is defined on attention maps created for both the teacher and student networks by summing over the channels of a feature map.\\

%\item Overhaul Distillation \cite{heo2019comprehensive}: The proposed method suggests distilling pre-ReLU information using a novel loss function designed for this purpose.\\

\item Similarity Preserving (SP)\cite{tung2019similarity}: The proposed method involves pair-wise transfer of information at the instance level between each pair of images in a batch.\\

\item Inter-Channel Knowledge Distillation (ICKD)\cite{liu2021exploring}: Explores inter-channel information by distilling inter-channel matrices between the teacher and student networks.\\
\end{itemize}
\end{comment}

All of the aforementioned methods were tested on both intermediate layers and final output maps to find the best results. Table \ref{tab: pascal-results} shows that the proposed AICSD method outperforms all of the mentioned methods with different student architectures on the validation set of the Pascal VOC 2012 dataset. %Although the proposed method is computationally expensive in terms of training time, as calculating pair-wise KL distances is more expensive than using inner products, the computation is only required during training.

Table \ref{tab: cityscapes-results} represents a comparison between the proposed method and other distillation methods in terms of mIoU and pixel accuracy on the validation set of the Cityscapes dataset. Although the degree of improvement achieved by the proposed method varies depending on the student network architecture, it consistently outperforms the other methods across different student backbones.

Moreover, Figure \ref{fig:moiu_classwise} compares the per-class mIoU of our proposed AICSD with KD and ICKD methods on the Cityscapes dataset. As shown in this figure, proposed method achieves significant improvements on some classes while maintaining similar performance on other classes. This suggests that the proposed method is able to capture and transfer new information from the teacher to the student that other methods may not capture.

\input{figures/visualization}

\subsection{Ablation Studies}
As introduced in the previous section, the proposed method includes the ICSD loss and the known KD loss, which are used together with the ALW training strategy. Table \ref{tab: ablation} presents an analysis of the impact of each of these methods on the improvement of the student network. The experiments conducted on the Pascal VOC 2012 dataset with two different backbone architectures demonstrate that the ALW strategy results in better performance than separately using the KD or ICSD losses. Moreover, with the MobileNet backbone, it can improve both the KD and ICSD methods by approximately 1.5 $\%$, in terms of mIoU.

Table \ref{tab: ALW-ablation} compares two different variants of the ALW training strategy, namely linear and exponential loss weighting, and demonstrates that both methods can improve the results of the student network. Additionally, depending on the architecture of the student network, one method may achieve better results than the other.

\input{tables/ablation}

\input{tables/ALW-ablation}

\subsection{\textit{Qualitative Assessment}}
To further validate the effectiveness of the proposed method, some qualitative assessments of the model’s performance are conducted. Figure \ref{fig:ablation1} visualizes the intra-class distributions and inter-class similarity matrices for a given image in the Cityscapes and Pascal VOC datasets. The top rows show two images from the Cityscapes dataset, their corresponding pair-wise matrices, and the intra-class distributions of road, rider, bus, and motorcycle classes (from top left to bottom right). The bottom rows show the same for images from the Pascal VOC dataset and the intra-class distributions of sheep, person, dog, and car classes. The figure compares the results of student network without distillation, student network trained with our proposed AICSD method, and the teacher network. The qualitative results show that our proposed method helps the student preserve the structure from the teacher and leads to both intra-class distributions and pair-wise matrices that are more similar to the teacher.

Figure \ref{fig:visualization}, on the other hand, represents output masks for KD, ICKD, and our proposed AICSD method on the Cityscapes and Pascal VOC datasets. As can be seen from this figure, and validated by Figure \ref{fig:moiu_classwise}, the two other distillation methods have poor performance for some classes (such as train and bus). In contrast, the proposed method can address this issue and improve the results for these two classes. The same improvement is observed for images from the Pascal VOC dataset, where the proposed method improves the results of classes such as bicycle, boat, and sofa by a good margin compared to the two other distillation methods.

%In conclusion, the experiments conducted in this research demonstrate that inter-class similarities have good potential for distillation as a pair-wise method. Moreover, adaptively changing the scale of both CE loss and distillation losses can significantly improve the performance of the network. The ALW strategy can be combined with any pair-wise distillation method to achieve better results.

\input{figures/ablation1}

%% file: figures/miou_classwise.tex
\begin{figure*}[!ht]
\centerline{\includegraphics[scale=0.74]{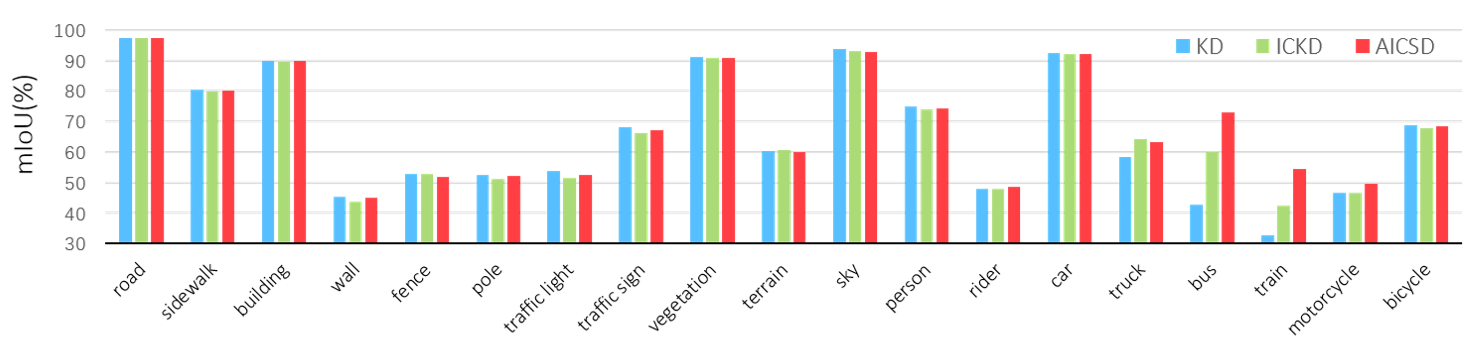}}
\caption{Comparison of mIoU per class between KD, ICKD, and proposed AICSD on the validation set of Cityscapes dataset. Student network uses a ResNet18 backbone.}
\label{fig:moiu_classwise}
\end{figure*}

%% file: tables/pascal-results.tex
\begin{table}[ht]
    \caption{\label{tab: pascal-results}Performance comparison of AICSD with other distillation methods for two different backbones on Pascal VOC 2012 Validation set.}
    \centering
    \begin{tabular}{l c c}
        \toprule
        Method &  mIoU(\%) & Params(M) \\
        \midrule 
        Teacher: Deeplab-V3 + (ResNet-101)  & 77.85  & 59.3 \\
        Student1: Deeplab-V3 + (ResNet-18) & 67.50 & 16.6 \\
        Student2: Deeplab-V3 + (MobileNet-V2) & 63.92 & 5.9 \\
        \midrule
        Student1 + KD  & 69.13 $\pm$  0.11  & 16.6 \\
        Student1 + AT  & 68.95 $\pm$ 0.26 & 16.6 \\
        Student1 + SP  & 69.04 $\pm$ 0.10 & 16.6 \\
        %Student1 + Overhaul  & 68.47 $\pm$ 0.10  & 16.6 \\
        Student1 + ICKD  & 69.13 $\pm$ 0.17 & 16.6 \\
        Student1 + AICSD (ours)   & \textbf{70.03 $\pm$ 0.13} & 16.6 \\
        \midrule
        Student2 + KD  & 66.39 $\pm$  0.21  & 5.9 \\
        Student2 + AT  & 66.27 $\pm$ 0.17 & 5.9 \\
        Student2 + SP  & 66.32 $\pm$ 0.05 & 5.9 \\
        %Student2 + Overhaul  & 66.47 $\pm$ 0.10  & 5.9 \\
        Student2 + ICKD  & 67.01 $\pm$ 0.10 & 5.9 \\
        Student2 + AICSD (ours)  & \textbf{68.05 $\pm$ 0.24} & 5.9 \\
        \bottomrule 
    \end{tabular}
\end{table}

%% file: tables/cityscapes-results.tex
\begin{table}[ht]
    \caption{\label{tab: cityscapes-results}Performance comparison on Cityscapes Validation set.}
    \centering
    \begin{tabular}{l c c}
        \toprule
        Method &  mIoU(\%) & Accuracy(\%) \\
        \midrule 
        T: ResNet101  & 77.66  & 84.05 \\
        S1: ResNet18 & 64.09 & 74.8 \\
        S2: MobileNet v2 & 63.05 & 73.38 \\
        \midrule
        S1 + KD  & 65.21 \textcolor{black}{(+1.12)} & 76.32 \textcolor{black}{(+1.74)} \\
        S1 + AT  & 65.29 \textcolor{black}{(+1.20)}& 76.27 \textcolor{black}{(+1.69)} \\
        S1 + SP  & 65.64 \textcolor{black}{(+1.55)} & 76.90 \textcolor{black}{(+2.05)} \\
        %S1 + Overhaul  & 68.47 & 16.6 \\
        S1 + ICKD  & 66.98 \textcolor{black}{(+2.89)} & 77.48 \textcolor{black}{(+2.90)} \\
        S1 + AICSD (ours)   &  \textbf{68.46 (+4.37)} &  \textbf{78.30 (+3.72)} \\
        \midrule
        S2 + KD  & 64.03 \textcolor{black}{(+0.98)} & 75.34 \textcolor{black}{(+1.96)} \\
        S2 + AT  & 63.72 \textcolor{black}{(+0.67)}& 74.79 \textcolor{black}{(+1.41)} \\
        S2 + SP  & 64.22 \textcolor{black}{(+1.17)} & 75.28 \textcolor{black}{(+1.90)} \\
        %S2 + Overhaul  & 66.47 & 5.9 \\
        S2 + ICKD  & 65.55 \textcolor{black}{(+2.50)} & 76.48 \textcolor{black}{(+3.10)} \\
        S2 + AICSD (ours)  &  \textbf{66.53 (+3.48)} &  \textbf{76.96 (+3.58)} \\
        \bottomrule 
    \end{tabular}
\end{table}

%% file: figures/visualization.tex
%\begin{figure}[!ht]
%\centering
%\centerline{\includegraphics[width=8.75cm, height=12cm]{images/visualization.png}}
%\caption{Qualitative comparison of results on Cityscapes and Pascal VOC validation sets. First two columns show results on Cityscapes and last two columns show results on Pascal VOC. Performance improvement of proposed AICSD method is compared with KD and ICKD methods.}
%\label{fig:visualization}
%\end{figure}

\begin{figure*}[!ht]
\centering
\centerline{\includegraphics[scale=1.05]{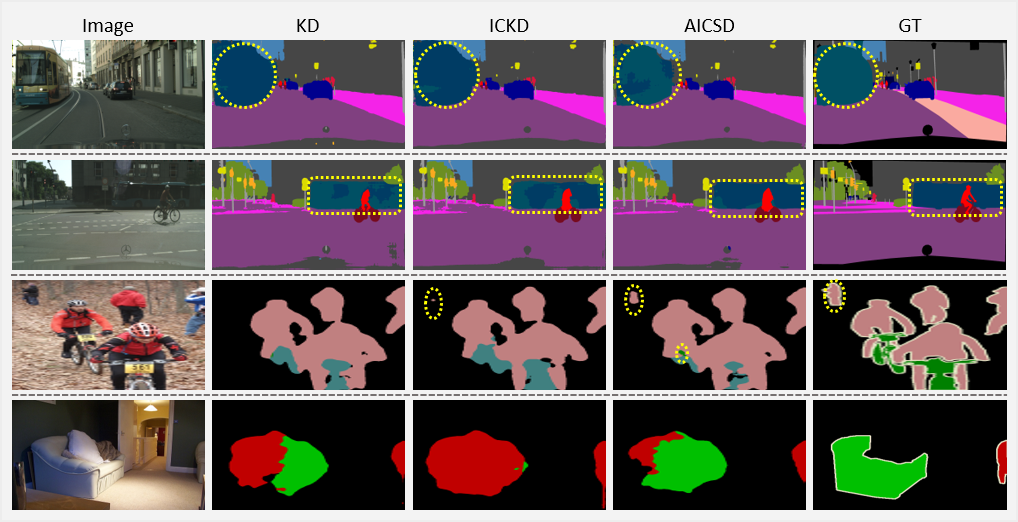}}
\caption{Qualitative comparison of results on Cityscapes and Pascal VOC validation sets. First two rows show results on Cityscapes and last two rows show results on Pascal VOC. Performance improvement of proposed AICSD method is compared with KD and ICKD methods.}
\label{fig:visualization}
\end{figure*}

%% file: tables/ablation.tex
\begin{table}[ht]
    \caption{\label{tab: ablation}Ablation study on PASCAL VOC 2012. Validating effectiveness of proposed ICSD and ALW training strategy.}
    \centering
    \begin{tabular}{l c c c c}
        \toprule
        Method &  KD & ICSD  & ALW  &  \textit{val} mIoU(\%) \\
        \midrule 
        Teacher:ResNet101  &    &  &    & 77.85\\

        \midrule
        Student1:ResNet18  &  n/a  & n/a &  n/a  & 67.50\\
        Student1:ResNet18  &  \checkmark  & \xmark &  \xmark  & 69.13 \textcolor{red}{(+1.63)}\\
        Student1:ResNet18  &  \xmark  & \checkmark &  \xmark  & 69.20 \textcolor{red}{(+1.70)}\\
        Student1:ResNet18  &  \checkmark  & \xmark &  \checkmark  & 69.48 \textcolor{red}{(+1.94)}\\
        Student1:ResNet18  &  \checkmark  & \checkmark &  \checkmark  & 70.03 \textcolor{red}{(+2.53)}\\
        \midrule
        Student2:MbileNetV2  &  n/a  & n/a &  n/a  & 63.92\\
        Student2:MbileNetV2  &  \checkmark  & \xmark &  \xmark  & 66.39 \textcolor{red}{(+2.47)}\\
        Student2:MbileNetV2  &  \xmark  & \checkmark &  \xmark  & 66.58 \textcolor{red}{(+2.66)}\\
        Student2:MbileNetV2  &  \checkmark  & \xmark &  \checkmark  & 67.02 \textcolor{red}{(+3.10)}\\
        Student2:MbileNetV2  &  \checkmark  & \checkmark &  \checkmark  & 68.05 \textcolor{red}{(+4.13)}\\
        \bottomrule 
    \end{tabular}
\end{table}

%% file: tables/ALW-ablation.tex
\begin{table}[htbp]
\caption{Ablation  for proposed ALW training strategy for two different approaches, Linear and Exponential, on PASCAL VOC 2012 Validation set.}
\begin{center}
\begin{tabular}{  l | c  c }
    \hline
    Method & Linear ALW & Exponential ALW   \\ \hline
    
    {S1: ResNet18} & 69.74 \textcolor{red}{(+2.24)} & 70.03 \textcolor{red}{(+2.53)}\\
    {S2: MobileNetV2} & 68.05 \textcolor{red}{(+4.13)} & 67.78 \textcolor{red}{(+3.87)} \\ 
    
    \hline
    \end{tabular}
\label{tab: ALW-ablation}
\end{center}
\end{table}

%% file: figures/ablation1.tex
\begin{figure*}[!ht]
\centerline{\includegraphics[scale=.64]{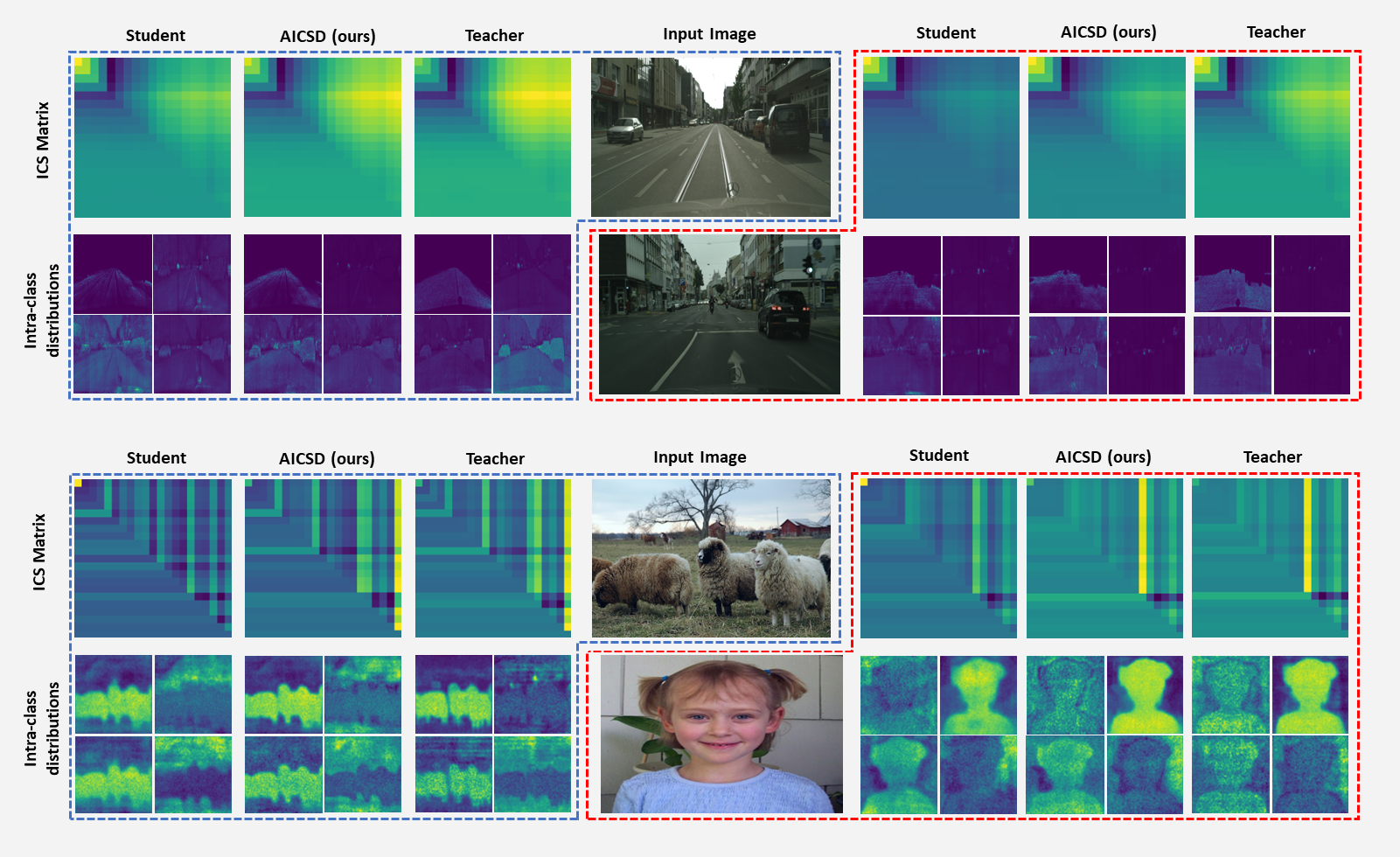}}
\caption{Illustration of the intra-class distributions and inter-class similarity matrices for selected images, from the validation sets of Cityscapes and Pascal VOC datasets. For each image, the top row shows the inter-class similarity matrix and the bottom row shows the intra-class distributions of four classes. Proposed distillation method trains the student network to mimic the structures of the teacher network. Student backbone is MobileNet for Pascal VOC dataset and ResNet18 for Cityscapes images.}
\label{fig:ablation1}
\end{figure*}

%% file: docs/5_conclusion.tex
\label{sec: conclusion}
This paper pressents the effective usage of knowledge distillation strategy to improve the performance of a lightweght network with the help of a larger and more complex network. In addition to the pixel-wise knowledge distillation method, the paper introduces a pair-wise method and a training strategy to enhance the knowledge distillation process. The proposed pair-wise method enhances the results of the student network by transferring inter-class similarities created from the outputs of the networks, making it appliable to  any semantic segmentation network architecture. The training strategy also boosts results by combining the proposed ICSD method with the pixel-wise distillation in an adaptive manner that weights the losses that control the model.

Extensive experiments were conducted on two challenging datasets, using two different student networks, to validate the effectiveness of the proposed method. Ablation studies were also performed to demonstrate the impact of the ALW strategy, which can be combined with other distillation methods as well.